\documentclass[10pt,twocolumn,letterpaper]{article}

\usepackage{cvpr}              

\usepackage{graphicx}
\usepackage{amsmath}
\usepackage{amssymb}
\usepackage{booktabs}

\usepackage{multirow}
\usepackage{caption}
\usepackage{sidecap}
\usepackage{wrapfig}
\usepackage{tabularx, booktabs, ragged2e}
\usepackage{adjustbox}
\usepackage[table]{xcolor}

\makeatletter
\newcommand*\bigcdot{\mathpalette\bigcdot@{.7}}
\newcommand*\bigcdot@[2]{\mathbin{\vcenter{\hbox{\scalebox{#2}{$\m@th#1\bullet$}}}}}
\makeatother

\usepackage[pagebackref,breaklinks,colorlinks]{hyperref}

\usepackage[capitalize]{cleveref}
\crefname{section}{Sec.}{Secs.}
\Crefname{section}{Section}{Sections}
\Crefname{table}{Table}{Tables}
\crefname{table}{Tab.}{Tabs.}

\newcommand{\argmin}{\mathop{\mathrm{argmin}}\limits}

\def\cvprPaperID{2018} 
\def\confName{CVPR}
\def\confYear{2023}

\begin{document}

\title{NeRFInvertor: High Fidelity NeRF-GAN Inversion \\ for Single-shot Real Image Animation}

\author{Yu Yin$^{1}$, Kamran Ghasedi$^{2}$, HsiangTao Wu$^{2}$, Jiaolong Yang$^{2}$, Xin Tong$^{2}$, Yun Fu$^{1}$\\
{\tt\small $^{1}$ Northeastern University}\\
{\tt\small yin.yu1@northeastern.edu, yunfu@ece.neu.edu}\\
{\tt\small $^{2}$ Microsoft}\\
{\tt\small kamran.ghasedi@gmail.com, \{musclewu,jiaoyan,xtong\}@microsoft.com}
}

\maketitle

\begin{abstract}
   Nerf-based Generative models have shown impressive capacity in generating high-quality images with consistent 3D geometry. Despite successful synthesis of fake identity images randomly sampled from latent space, adopting these models for generating face images of real subjects is still a challenging task due to its so-called inversion issue. In this paper, we propose a universal method to surgically fine-tune these NeRF-GAN models in order to achieve high-fidelity animation of real subjects only by a single image. Given the optimized latent code for an out-of-domain real image, we employ 2D loss functions on the rendered image to reduce the identity gap. 
    Furthermore, our method leverages explicit and implicit 3D regularizations using the in-domain neighborhood samples around the optimized latent code to remove geometrical and visual artifacts. 
   Our experiments confirm the effectiveness of our method in realistic, high-fidelity, and 3D consistent animation of real faces on multiple NeRF-GAN models across different datasets.
\end{abstract}

\section{Introduction}\label{sec:introduction}

\begin{figure}[t]
  \centering
   \includegraphics[width=\linewidth]{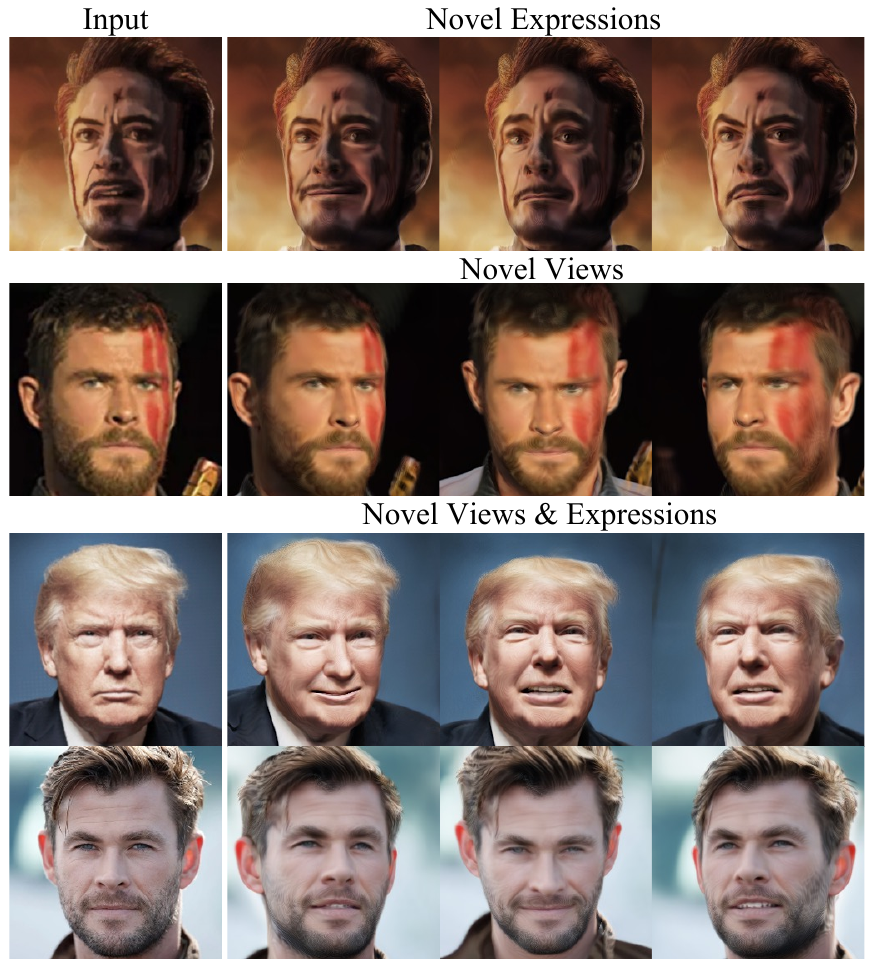}
   \caption{\textbf{Image animation results of our method.} \textit{NeRFInvertor} achieves 3D-consistent and ID-preserving animation (\ie novel views and expressions) of real subjects given only a single image.
   }
   \label{fig:teaser}
\end{figure}

Animating a human with a novel view and expression sequence from a single image opens the door to a wide range of creative applications, such as talking head synthesis~\cite{wang2021one,meshry2021learned}, augmented and virtual reality (AR/VR)~\cite{li20193d}, image manipulation~\cite{tov2021designing,park2020swapping}, as well as data augmentation for training of deep models~\cite{pranjal2019palmprint,liu2018semi}.
Early works of image animation mostly employed either 2D-based image generation models~\cite{geng2018warp,wu2021f3a,pumarola2020ganimation,tewari2020pie}, or 3D parametric models~\cite{danvevcek2022emoca,ye20203d,bao2021high,ye2022high} (\eg 3DMM~\cite{blanz1999morphable}), but they mostly suffer from artifacts, 3D inconsistencies or unrealistic visuals.

Representing scenes as Neural Radiance Fields (NeRF)~\cite{mildenhall2021nerf} has recently emerged as a breakthrough approach for generating high-quality images of a scene in novel views. However, the original NeRF models ~\cite{barron2021mip,yu2021plenoctrees,wang2022fourier} only synthesize images of a static scene and require extensive multi-view data for training, restricting its application to novel view synthesis from a single image.
Several studies have shown more recent advances in NeRFs by extending it to generate multi-view face images with single-shot data even with controllable expressions \cite{cai2022pix2nerf,zhang2022monocular,rebain2022lolnerf,wu2022anifacegan,deng2022gram,chan2022eg3d}. 
These Nerf-based Generative models (NeRF-GANs) are able to embed attributes of training samples into their latent variables, and synthesize new identity face images with different expressions and poses by sampling from their latent space.

\begin{figure}[t]
  \centering
  \includegraphics[width=\linewidth]{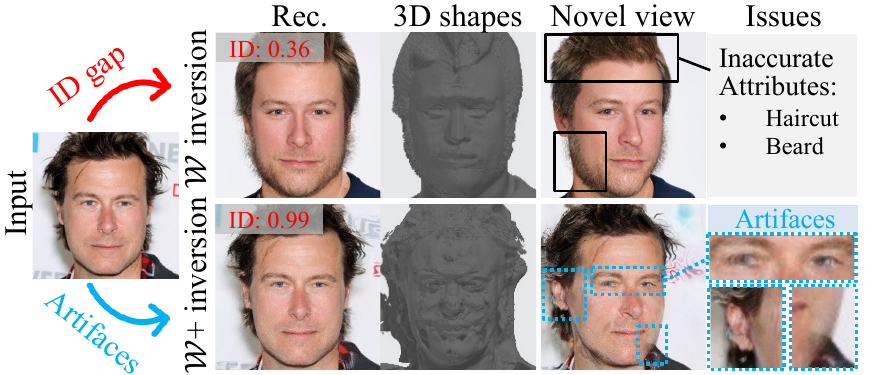}
   \caption{\textbf{Trade-off between ID-preserving and removing artifacts.} Optimizing latent variables of Nerf-GANs for synthesis of a real face leads to a trade-off between identity-preserving and geometrical and visual artifacts. Specifically, $\mathcal{W}$ space inversion results in clean geometry but identity gap between real and generated images, and $\mathcal{W+}$ space inversion causes preserving of identity attributes but inaccurate geometry and visual artifacts. 
   }
   \label{fig:id_geo_tradeoff}
\end{figure}

While animatable synthesis of fake identity images is impressive, it is still challenging to generate 3D-consistent and identity-preserving images of real faces.
Specifically, current Nerf-GANs have difficulties to accurately translate out-of-domain images into their latent space, and consequently change identity attributes and/or introduce artifacts when applied to most real-world images. 
In order to synthesize real faces, the conventional method applies optimization algorithms to invert the input image to a latent code in a smaller (\ie $\mathcal{W}$) or an extended (\ie $\mathcal{W+}$) NeRF-GAN latent space. However, they both either have ID-preserving or artifacts issues as shown in Figure~\ref{fig:id_geo_tradeoff}. The $\mathcal{W}$ space inversion, in particular, generates realistic novel views and clean 3D geometries, but suffers from the identity gap between the real and synthesized images. 
In contrast, the $\mathcal{W+}$ space inversion well preserves the identity but commonly generates an inaccurate 3D geometry, resulting in visual artifacts when exhibited from new viewpoints.
Hence, it remains as a trade off to have a 3D-consistent geometry or preserve identity attributes when inverting face images out of latent space distribution.

In this paper, we present \textit{NeRFInvertor} as a universal inversion method for NeRF-GAN models to achieve high-fidelity, 3D-consistent, and identity-preserving animation of real subjects given only a single image. 
Our method is applicable to most of NeRF-GANs trained for a \textit{static} or \textit{dynamic} scenes, and hence accomplish synthesis of real images with both novel views and novel expressions (see Figure~\ref{fig:teaser}).
Since the real images are mostly out of the domain of NeRF-GANs latent space, we surgically fine-tune their generator to enrich the latent space by leveraging the single input image  without degrading the learned geometries.

In particular, given an optimized latent code for the input image, we first use image space supervision to narrow the identity gap between the synthesized and input images. Without a doubt, the fine-tuned model can be overfitted on the input image and well reconstruct the input in the original view. However, fine-tuning with just image space supervision produces erroneous 3D geometry due to the insufficient geometry and content information in a single image, resulting in visual artifacts in novel views. To overcome this issue, we introduce regularizations using the surrounding samples in the latent space, providing crucial guidance for the unobserved part in the image space. By sampling latent codes from the neighborhood of optimized latent variables with different poses and expressions, we enforce a novel geometric constraint on the density outputs of fine-tuned and original pretrained generators.
We also further add regularizations on the rendered images of neighborhood samples obtained from the fine-tuned and pretrained generators. These regularizations help us to leverage the geometry and content information of those in-domain neighborhood samples around the input.
Our experiments validate the effectiveness of our method in realistic, high-fidelity, and 3D consistent animating of real face images.

The main contributions of this paper are as follows: 
\begin{enumerate}
    \item We proposed a universal method for inverting NeRF-GANs to achieve 3D-consistent, high-fidelity, and identity-preserving animation of real subjects given only a single image.
    
    \item We introduce a novel geometric constraint by leveraging density outputs of in-domain samples around the input to provide crucial guidance for the unobserved part in the 2D space.
    
    \item We demonstrate the effusiveness of our method on multiple NeRF-GAN models across different datasets.

\end{enumerate}

\section{Related Work}\label{sec:related_work}
\begin{figure*}[t]
  \centering
   \includegraphics[trim=0.31in 0in 0.19in 0in,clip,width=\linewidth]{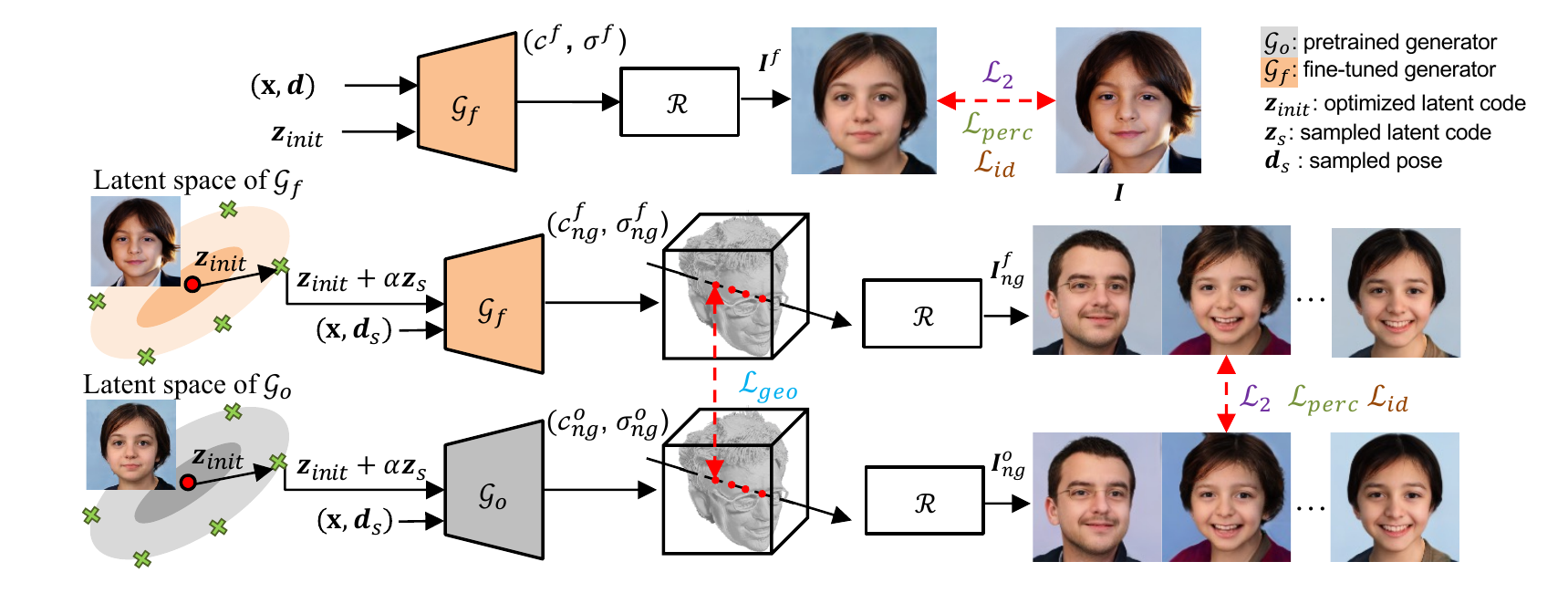}
   \caption{\textbf{Framework of \textit{NeRFInvertor}.} Given the optimized latent code $\textbf{z}_{init}$, we fine-tune the generator and first apply image space supervision to push the generated image to match the input image in the original view $\textbf{\textit{d}}$.
   To augment the NeRF-GAN manifold without worrying about visual artifacts in novel views, we then leverage the surrounding samples of the optimized latent code to regularize the realism and fidelity of the novel view and expression synthesis.
    }
   \label{fig:onecol}
\end{figure*}

\subsection{NeRF-GANs}
Recently, the impressive performance of NeRF-GANs has demonstrated its potential as a promising research direction. 
GRAF~\cite{schwarz2020graf} and Pi-GAN~\cite{chan2021pigan} are two early attempts that proposed generative models for radiance fields for 3D-aware image synthesis from the unstructured 2D images. 
Several recent studies (\eg GRAM~\cite{deng2022gram} and EG3D~\cite{chan2022eg3d}) have enhanced synthesis quality with higher resolutions, better 3D geometry, and faster rendering. 
Furthermore, AniFaceGAN~\cite{wu2022anifacegan} introduced a deformable NeRF-GAN for dynamic scenes capable of synthesizing faces with controllable pose and expression.
In this paper, we demonstrate the effectiveness of our method on NeRF-GANs for both static (GRAM and EG3D) and dynamic (AniFaceGAN) scenes.

\subsection{GAN Inversion}
GAN priors can be beneficial in a variety of applications such as image editing~\cite{abdal2021styleflow,collins2020editing,richardson2021encoding} and face restoration~\cite{yang2021gan,wang2021towards}.
To utilize the priors, the input image needs to be inverted into a latent code that optimally reconstructs the given image using a pretrained generator. The process is known as GAN inversion, and it is a well-known issue in 2D GANs (\eg StyleGAN~\cite{karras2019style,Karras2019stylegan2,Karras2020ada}).
In the 2D domain, Abdal \etal~\cite{abdal2019image2stylegan,abdal2020image2stylegan++} employed direct optimization approach to invert an image
to styleGAN's latent space and demonstrated the distortion-editability trade-off between $\mathcal{W}$ and $\mathcal{W+}$ space. PTI~\cite{roich2022pti} achieved promising performance by using an initial inverted latent code as a pivot and fine-tuning the generator to mitigate the distortion-editability trade-off.

To exploit NeRF-GAN priors for 3D tasks, recent attempts inverted the real image to NeRF-GAN latent space using the traditional optimization method~\cite{zhang2022monocular}, additive encoder~\cite{cai2022pix2nerf,rebain2022lolnerf}, or PTI~\cite{chan2022eg3d,lin20223d}. However, applying these 2D-based inversion methods directly to 3D-GAN models makes them insensitive to subtle geometric alterations in 3D space. 
To address this issue, we propose explicit geometrical regularization and implicit geometrical regularization to help produce better 3D geometry of the input.

\subsection{Single-shot NeRFs}
Due to the insufficient geometry and content information in a single image, single-shot NeRFs without additional supervision (\ie multi-view images or 3D objects) remain challenging. Recently, Pix2NeRF~\cite{cai2022pix2nerf} extended pi-GAN with the encoder to obtain a conditional single-shot NeRF models. HeadNeRF~\cite{hong2022headnerf} incorporated the NeRF into the parametric representation of the human head. Since the whole process is differentiable, the model can generate a NeRF representation from a single image using image fitting.
However, there remains a notable identity gap between the synthesized results and input images, indicating the inefficiency of single-shot NeRFs for real image inversion.

\section{Method}\label{sec:approach}

In this paper, we present \textit{NeRFInvertor} as a universal NeRF-GAN inversion method to translate a single real image into a NeRF representation. 
Given an input image $\textbf{\textit{I}}$, 
the goal is to generate novel views or expressions of $\textbf{\textit{I}}$ using a pretrained Nerf-GAN $\mathcal{G}_o$.
To do this, we first find an in-domain latent code\footnote{We optimize latent codes in $\mathcal{Z}$ or $\mathcal{W}$ space ~\cite{Karras2019stylegan2} for different NeRF-GAN models.} capable of generating an image $\textbf{\textit{I}}^o$ as close as possible to the input image.
Given a NeRF-GAN generator $\mathcal{G}_{(\cdot)}$, the image $\textbf{\textit{I}}^{(\cdot)}$ can be synthesized as:
\begin{equation}
\textbf{\textit{I}}^{(\cdot)} = \mathcal{R} (\mathcal{G}_{(\cdot)}(\mathbf{z}, \mathbf{x}), \textbf{\textit{d}}) \ ,
\end{equation}
where $\mathbf{x} \in \mathbb{R}^3$ is the a 3D location, $\mathbf{z}$ denotes a latent code, $\textbf{\textit{d}}\in \mathbb{R}^3$ is the camera pose, and $\mathcal{R}$ is the volume renderer discussed in \cite{max1995optical,mildenhall2021nerf}.
Therefore, the optimized latent code $\textbf{z}_{init}$ can be computed as:
\begin{equation}
\textbf{z}_{init} = \argmin_{\textbf{{z}}} \ \ \mathcal{L}_{perc}(\textbf{\textit{I}}, \textbf{\textit{I}}^o) + \lambda_0 \mathcal{L}_{pix}(\textbf{\textit{I}}, \textbf{\textit{I}}^o) \ , 
\end{equation} 
where $\mathcal{L}_{perc}$, $\mathcal{L}_{pix}$ represent perceptual and l2-norm pixel-wise loss functions, respectively.

Usually, there would be a gap between the generated and real images, since the real images are mostly the out-of-domain samples in NeRF-GANs. To solve this problem, we propose a fine-tuning process with novel regularizations in the following sections.
Specifically, we fine-tune the generator with image space loss functions (Sec. \ref{sec:2dloss}) to reduce the identity gap. We also apply an explicit geometrical constraint (Sec. \ref{sec:neiborhood_reg}) and an implicit geometrical regularization (Sec. \ref{sec:image_reg})  to maintain the model’s ability to produce high-quality and 3D-consistent images.

\subsection{Image Space Supervision}\label{sec:2dloss}
Given the optimized latent code $\textbf{z}_{init}$, we fine-tune the generator using image space supervision by pushing the generated image to match the input image in the original view $\textbf{\textit{d}}$.
Denoting the fine-tuned generator as $\mathcal{G}_f$, it takes the optimized latent code $\textbf{z}_{init}$ and a 3D location $\mathbf{x} \in \mathbb{R}^3$ as the inputs and outputs a color $c \in \mathbb{R}^3$ and a volume density $\sigma \in \mathbb{R}^1$ for each location. 
For the given view $\textbf{\textit{d}}$, we can then accumulate  the colors and densities into a 2D image $\textbf{\textit{I}}^f$. Formally, the image $\textbf{\textit{I}}^f$ can be expressed as:

\begin{equation}
\textbf{\textit{I}}^f = \mathcal{R} (\mathcal{G}_f(\mathbf{z}_{init}, \mathbf{x}), \textbf{\textit{d}})
\end{equation}

We employ the following loss function as the image space supervision:
\begin{equation}
\mathcal{L}_{img} = \lambda_{1} \mathcal{L}_{pix}(\textbf{\textit{I}}^{f}, \textbf{\textit{I}}) + \lambda_2 \mathcal{L}_{perc}(\textbf{\textit{I}}^{f}, \textbf{\textit{I}}) + \lambda_{3} \mathcal{L}_{id}(\textbf{\textit{I}}^{f}, \textbf{\textit{I}}) \ ,
\end{equation}
where $\mathcal{L}_{perc}$, $\mathcal{L}_{pix}$ and $\mathcal{L}_{id}$ indicate perceptual, l2-norm pixel-wise and identity losses, respectively.
$\lambda_1$, $\lambda_2$ and $\lambda_3$ are hyper-parameters of the losses. 
With the image space supervision, the fine-tuned model well reconstructs the input in the original view, but is prone to overfitting on the input image,  causing artifacts in novel view synthesized images and inaccurate  3D geometry of the subject.

\begin{figure}[t]
  \centering
   \includegraphics[width=\linewidth]{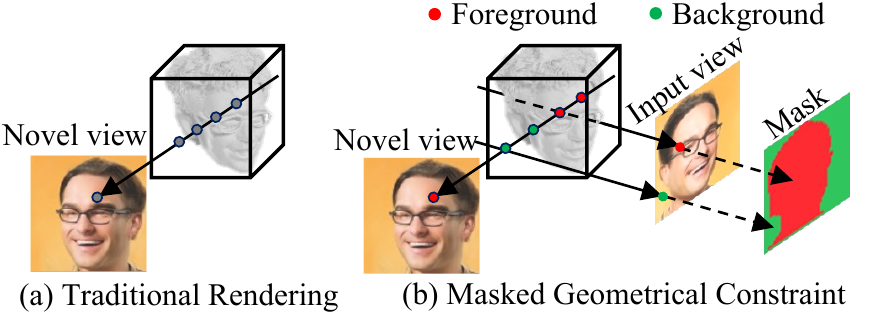}
   \caption{\textbf{Masked Geometrical Constraint.} (a) It shows the traditional rendering process for a novel view image. (b) Masked geometrical constraint takes only the foreground points into account. To render the novel view, only the red dots are used for image rendering. The two green dots are ignored, as they are definitely from a background region.
}
   \label{fig:masked_matting}
\end{figure}

\subsection{Explicit Geometrical Regularization}\label{sec:neiborhood_reg}

To enrich the NeRF-GANs manifold using the input image attributes without worrying about visual artifacts in novel views, we relax the assumption of strict image space alignment described in Sec. \ref{sec:2dloss}. In order to regularize the model, we leverage the neighborhood samples around the optimized latent code to enhance the geometry, realism, and fidelity of the novel view and expression synthesis.

We first randomly sample different neighborhood latent codes $\mathbf{z}_{ng}$ surrounding the optimized latent code with various poses and expressions. 
The neighborhood latent codes can be obtained by:
\begin{equation}
\textbf{z}_{ng} = \textbf{z}_{init} + \alpha \frac{\textbf{z}_{smp}-\textbf{z}_{init} }{||\textbf{z}_{smp}-\textbf{z}_{init} ||_2} \ ,
\end{equation}
where $\alpha$ is the interpolation distance between the randomly sampled $\textbf{z}_{smp} \sim \mathcal{N}(\textbf{0},\,\textbf{1})$ and optimized latent variable $\textbf{z}_{init}$.
To leverage the high-fidelity qualities of the original generator $\mathcal{G}_o$, we force the fine-tuned generator $\mathcal{G}_f$ to perform the same as $\mathcal{G}_o$ on the neighborhood latent codes, poses, and expressions. The geometrical constraint is defined based on both the color and density outputs of neighborhood samples on $\mathcal{G}_f$ and $\mathcal{G}_o$, that are expressed as: 
\begin{equation}
\begin{aligned}
c_{ng}^f, \sigma_{ng}^f = & \mathcal{G}_f(\mathbf{z}_{ng}, \textbf{x}), \\
c_{ng}^o, \sigma_{ng}^o = & \mathcal{G}_o(\mathbf{z}_{ng}, \textbf{x}),
\end{aligned}
\end{equation}
By reprojecting each ray (\emph{i.e.} pixel) to 3D space according to its depth, we define two sets of point clouds $S_o$ and $S_f$ based on $\mathcal{G}_o$ and $\mathcal{G}_f$ images. 
We compare the similarity of two point clouds using the Chamfer distance, and
define the geometrical constraint as follows:
\begin{equation}
\begin{aligned}
\mathcal{L}_{exp} = \frac{1}{|S_o|} \sum_{p_o \in S_o} min_{p_f \in S_f} &|| \sigma_{ng}^f(p_f) - \sigma_{ng}^o(p_o)||^2_2 + \\
&|| c_{ng}^f(p_f) - c_{ng}^o(p_o)||^2_2 
\end{aligned}
\end{equation}
where $p_o$ and $p_f$ are the 3D locations in the point cloud sets $S_o$ and $S_f$.

\begin{figure}[t]
  \centering
   \includegraphics[trim=0.2in 0.19in 0.2in 0in,clip,width=\linewidth]{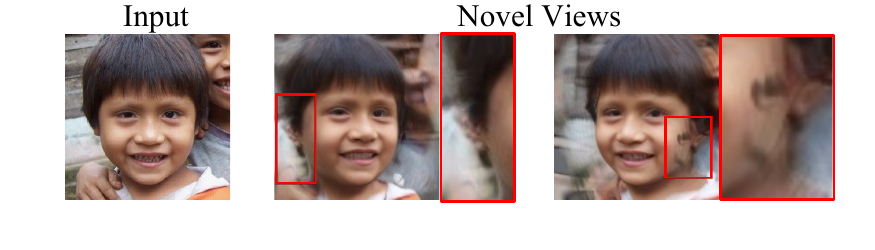}
   \caption{\textbf{The fogging artifacts around the hairline and/or in cheek region.}
}
   \label{fig:matting}
\end{figure}

\begin{figure*}[t]
  \centering
   \includegraphics[width=\linewidth]{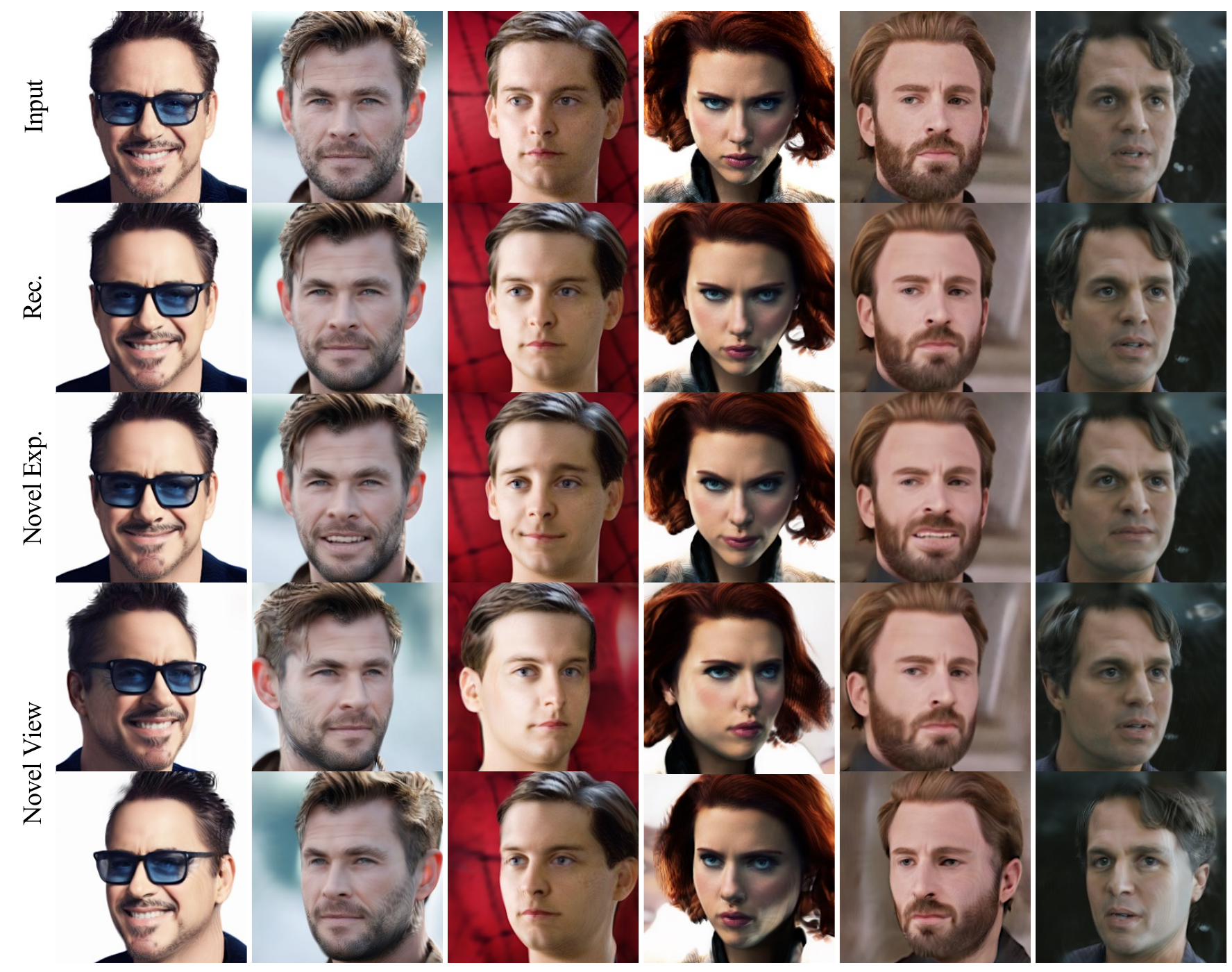}
   \caption{\textbf{Real image animation example on the ``Avengers''.} \textit{NeRFInvertor} synthesizes realistic faces using the pretrained AniFaceGAN with controllable pose and expression sequences given a single training image. As can be seen, \textit{NeRFInvertor} not only is capable of preserving identity attributes but also generates images with high quality and consistent appearance across different poses and expressions.
}
   \label{fig:showcase}
\end{figure*}

\subsection{Implicit Geometrical Regularization}\label{sec:image_reg}
Moreover, we also add implicit geometrical regularizations on the rendering results of fine-tuned and the pretrained generator. 
Given a novel view $\textbf{\textit{d}}_{s}$, the rendered image of $\mathcal{G}_f$ and  $\mathcal{G}_o$ can be expressed as:
\begin{equation}
\begin{aligned}
\textbf{\textit{I}}_{ng}^f = &\mathcal{R}((c_{ng}^f, \sigma_{ng}^f), \textbf{\textit{d}}_{s}), \\
\textbf{\textit{I}}_{ng}^o = &\mathcal{R}((c_{ng}^o, \sigma_{ng}^o), \textbf{\textit{d}}_{s}).
\end{aligned}
\end{equation}

We minimize the distance between the image generated by $\mathcal{G}_f$ and $\mathcal{G}_o$ using pixel-wise, perceptual, and identity losses. Hence, the overall loss can be expressed as:
\begin{equation}
\begin{aligned}
\mathcal{L} = & \mathcal{L}_{img} + \lambda_4 \mathcal{L}_{exp} + \lambda_5  \mathcal{L}_{pix}(\textbf{\textit{I}}_{ng}^f, \textbf{\textit{I}}_{ng}^o) + \\ &\lambda_6 \mathcal{L}_{perc}(\textbf{\textit{I}}_{ng}^f, \textbf{\textit{I}}_{ng}^o) + \lambda_7 \mathcal{L}_{id}(\textbf{\textit{I}}_{ng}^f, \textbf{\textit{I}}_{ng}^o),
\end{aligned}
\end{equation}

\subsection{Masked Regularizations}

Given a single-view image, we fine-tune the model with novel regularizations to achieve better 3D geometry and higher fidelity images in novel views. 
However, we still noticed some fogging parts around the hair or cheek as shown in Figure \ref{fig:matting}.
In order to remove artifacts and get more accurate geometry, We enhance our geometrical and image regularizations by a mask, which is based on matting information on the input image. 
As shown in Figure \ref{fig:masked_matting}, we predict the mask for the input view image based on the foreground and background regions shown by the red and green colors.
In particular, if a shooting ray reaches the foreground region of the input image, we classify all the sample points on that ray as foreground points.
Similarly, if a shooting ray reaches the background region of the image, we classify all the data points on that ray as background points.
In the masked constraints, we take only the foreground (red) points into account for neighborhood density, color, and image rendering, and ignore the background (green) points.

\section{Experiments}\label{sec:experiments}
In this section, we compare our approach to existing inversion methods as well as single-shot NeRF models qualitatively and quantitatively.
We validate our method on multiple NeRF-GANs trained for static (\ie GRAM and EG3D) or dynamic scene (\ie AniFaceGAN). Evaluations are performed on reconstruction, novel view, and expression synthesis. Finally, we conduct ablation studies to disclose the contributions of the proposed components of our work.

\begin{figure}[t]
  \centering
  \includegraphics[width=\linewidth]{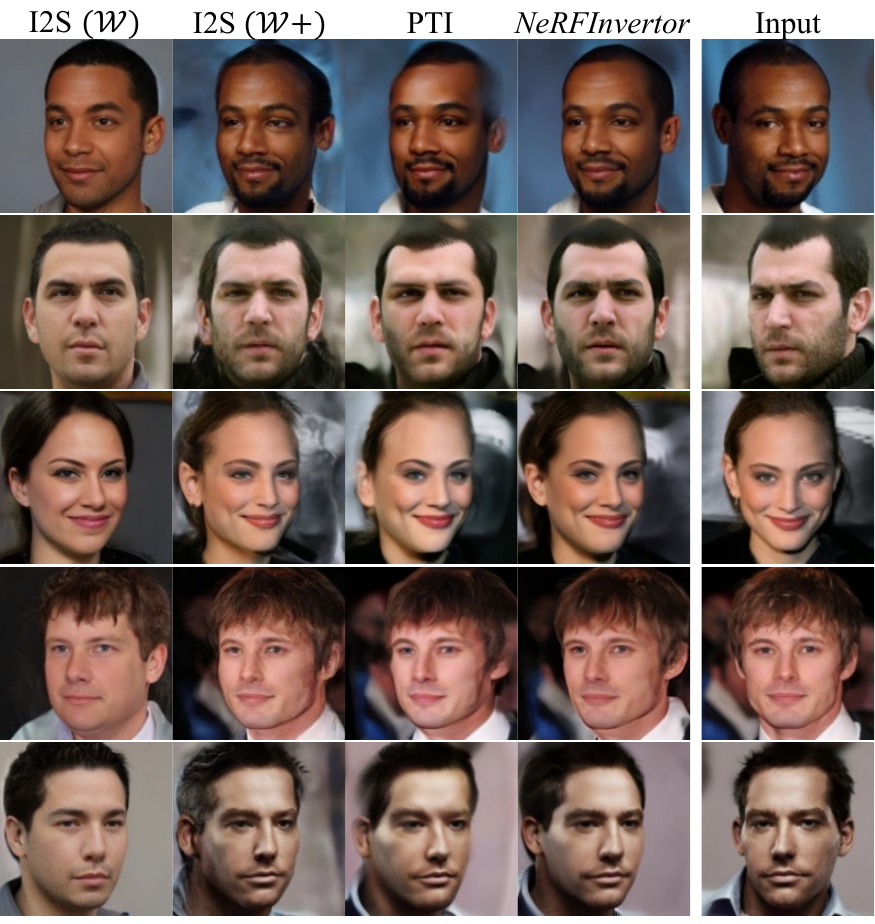}
   \caption{\textbf{Comparison with prior inversion methods.}
   I2S ($\mathcal{W}$) fails to preserve the identity of the input image, while I2S ($\mathcal{W+}$) introduces artifacts in novel views. PTI may generate texture that is inconsistent with the input image, such as fogging over a portion of the hair and ear or mismatched haircut with the input (see 2$^{nd}$ row). In comparison, our \textit{NeRFInvertor} delivers superior visual quality while preserving identity attributes.
   }
\label{fig:inversion_comparison}
\end{figure}

\subsection{Implementation Details}
We use AniFaceGAN generator by default for the experiments unless otherwise specified.  For face animation, we employ a generator pretrained on FFHQ dataset ~\cite{karras2019style}, and evaluate on both FFHQ and CelebA-HQ ~\cite{karras2017progressive} datasets. 
We also use a generator pretrained on Cats~\cite{zhang2008cat} to synthesize cat faces in novel views.
In addition, we collect images of famous people as out-of-domain samples to emphasize the \textit{NeRFInvertor}'s identity-preserving capability (Figure~\ref{fig:showcase}).

We describe the training details of our \textit{NeRFInvertor} applied on GRAM and EG3D in the Supplementary Material. 
For AniFaceGAN, the model is fine-tuned for 500 iterations. Hyper-parameters were set as follows: $\alpha=5$, $\lambda_{0}=0.1$, $\lambda_{1}=1$, $\lambda_{2}=10$, $\lambda_3=0.1$, $\lambda_4=10$, $\lambda_5=1$, $\lambda_6=10$, $\lambda_7=0.1$. The model is trained on 2 Nvidia RTX GPUs at the resolution of $128 \times 128$. We used an ADAM optimizer with a learning rate of $2e^{-5}$. Our model takes approximately 30 minutes for training.

\begin{figure}[t]
  \centering
  \includegraphics[width=0.85\linewidth]{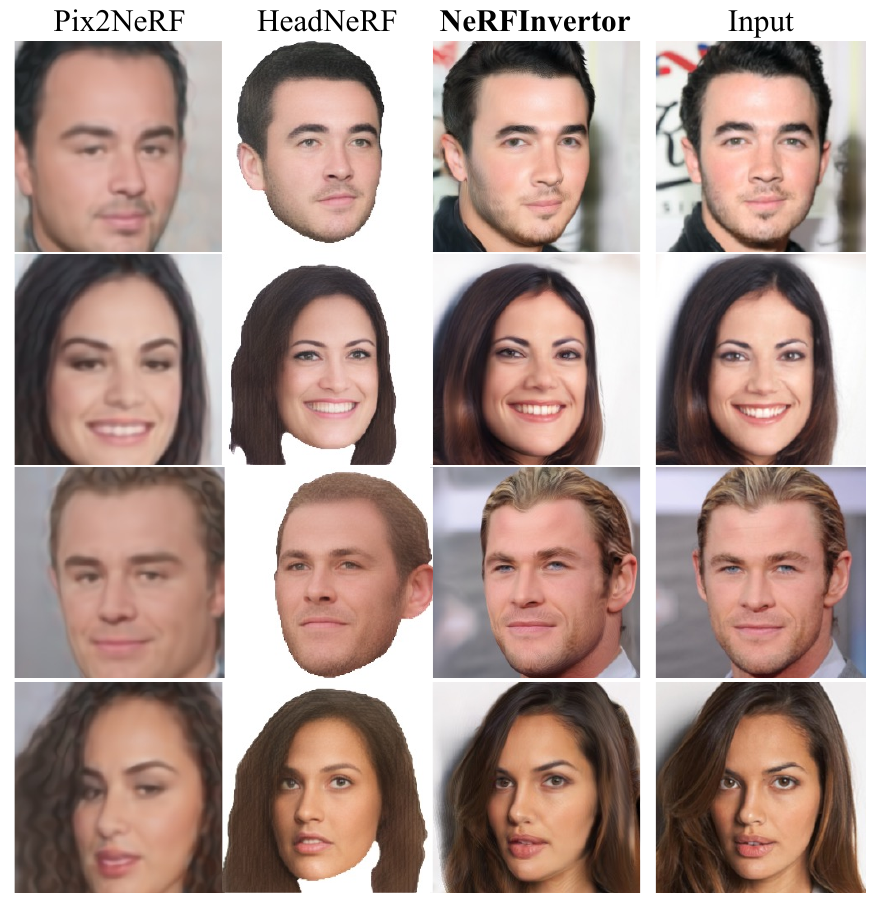}
   \caption{\textbf{Comparison with single-shot NeRF methods.}
   Existing single-shot NeRF approaches are either incapable of dealing with out-of-domain images or provide slightly artificial graphics. Our approach achieves 3D-consistent and ID-preserving animations.
   }
\label{fig:img2nerf_comparison}
\end{figure}

\begin{figure*}[t]
  \centering
  \includegraphics[width=\linewidth]{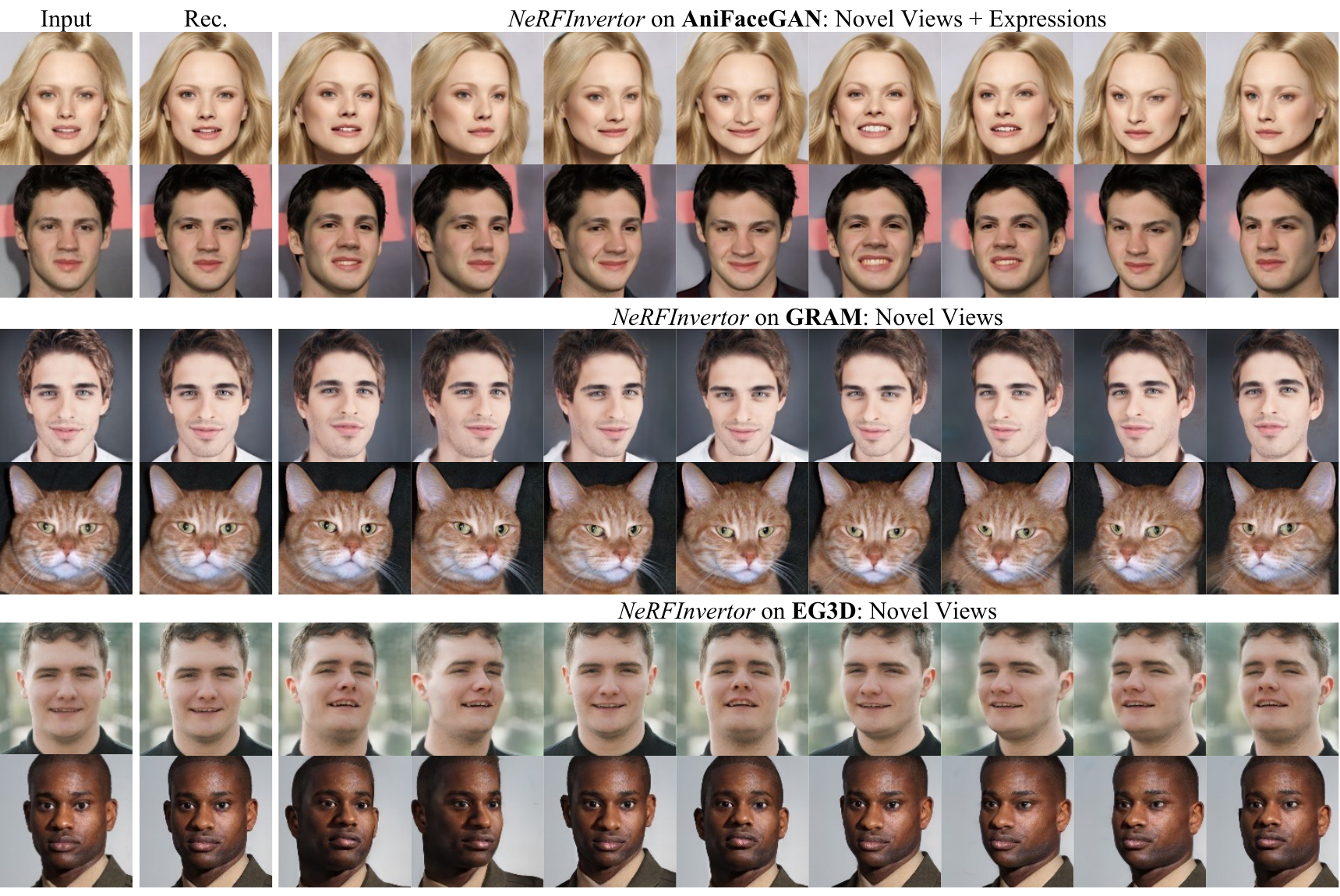}
   \caption{\textbf{Applying \textit{NeRFInvertor} on multiple NeRF-GANs.} We show reconstruction, novel views and expressions synthesis. \textit{NeRFInvertor} achieves high-fidelity, texture and 3D consistencies and ID-preserving synthesis across poses and expressions.}
   \label{fig:novel_views_from_multi_GAN}
\end{figure*}

\subsection{Qualitative Evaluation}
\noindent\textbf{Comparison with inversion methods.}
We start by qualitatively comparing our approach to prior inversion methods including I2S ($\mathcal{W}$), and I2S ($\mathcal{W+}$) and PTI. I2S~\cite{abdal2020image2stylegan++} employs the conventional optimization strategy to invert real images to either the smaller $\mathcal{W}$ space or the extended $\mathcal{W+}$ space. 
PTI~\cite{roich2022pti} is an inversion method for 2D-GANs with excellent performance. 
Figure~\ref{fig:inversion_comparison} shows a qualitative comparison of these methods for novel view synthesis from a single input image. I2S ($\mathcal{W}$) is able to generate images with acceptable visual quality, however there is a notable identity gap with the input image. In contrast, I2S ($\mathcal{W+}$) keeps identity attributes but introduces lots of artifacts in novel views. PTI shows better performance than I2S ($\mathcal{W}$) and I2S ($\mathcal{W+}$) by fine-tuning the model with image space supervisions. However, since it lacks explicit constraints in 3D space, it fails to generate accurate geometry of the input subject and the hidden content in the original view.
Our \textit{NeRFInvertor} outperforms these methods in terms of visual quality and identity preservation. 
Furthermore, our method can generate novel view images with similar texture to the input image. For example, the visual details in Figure~\ref{fig:inversion_comparison} such as wrinkle on the forehead (1$^{st}$ row) and dimples in the cheeks (3$^{rd}$ row) are well maintained in novel views.
In the Supplementary Material, we demonstrate reconstruction results and show more novel views and expressions synthesis.

\begin{table}
    \centering
    \caption{\textbf{Quantitative comparisons on FFHQ and CelebA-HQ test set.} The \textbf{best}, and the {\color{blue} second best} scores are highlighted.
    }
    {
    \begin{adjustbox}{max width=\linewidth}
        \begin{tabular}{p{10mm}cccc|cccc}
            \toprule
             && \multicolumn{3}{c|}{FFHQ} & \multicolumn{3}{c}{CelebA-HQ} \\
            NeRF- & Inversion & \multicolumn{1}{c}{Rec.} & \multicolumn{2}{c|}{Novel View} & \multicolumn{1}{c}{Rec.} & \multicolumn{2}{c}{Novel View}\\
            GANs & Methods & PSNR ($\uparrow$) & FID ($\downarrow$) & ID ($\uparrow$) & PSNR ($\uparrow$) & FID ($\downarrow$) & ID ($\uparrow$)\\\midrule 
            
            {\multirow{4}{*}{\shortstack{AniFace\\\cite{wu2022anifacegan}} }} & I2S~\cite{abdal2020image2stylegan++} ($\mathcal{W}$) & 15.62 & 58.37 & 0.33 & 16.50 & 42.53 & 0.25 \\
            \multirow{4}{*}{} & I2S~\cite{abdal2020image2stylegan++} ($\mathcal{W+}$) & \textbf{25.44} & 65.36 & \textbf{0.83} & \textbf{26.86} & 45.08 & {\color{blue} 0.76} \\
            \multirow{4}{*}{} & PTI~\cite{roich2022pti} & 23.89 & {\color{blue} 50.20} & 0.75 & 22.61 &  {\color{blue} 37.83} & 0.73 \\ \rowcolor{lightgray}
            \multirow{4}{*}{} &  \textit{NeRFInvertor} & {\color{blue} 24.92} & \textbf{45.89} & {\color{blue} 0.76} & {\color{blue} 25.61} & \textbf{34.07} & \textbf{0.77} \\
            \midrule 
            
            {\multirow{4}{*}{\shortstack{GRAM\\\cite{deng2022gram}}}} & I2S~\cite{abdal2020image2stylegan++} ($\mathcal{W}$) & 16.74 & 49.65 & 0.39 & 17.57 & \textbf{30.95} & 0.19 \\
            \multirow{4}{*}{} & I2S~\cite{abdal2020image2stylegan++} ($\mathcal{W+}$)  & 26.98 & 64.97 & 0.66 & 27.61 & 46.92 & 0.70 \\
            \multirow{4}{*}{} & PTI~\cite{roich2022pti}  & \textbf{28.90} & {\color{blue} 45.94} & {\color{blue} 0.79} & \textbf{29.26} & 38.01 & {\color{blue} 0.80} \\  \rowcolor{lightgray}
            \multirow{4}{*}{} & \textit{NeRFInvertor} & {\color{blue} 28.46} & \textbf{43.58} & \textbf{0.80} & {\color{blue} 28.75} & {\color{blue} 31.11} & \textbf{0.81} \\
            \bottomrule
        \end{tabular}
    \label{tab:qualitative_comparison}
    \end{adjustbox}
    }
\end{table}

\noindent\textbf{Comparison with single-shot NeRF methods.}
We compare our method with single-image view synthesis methods, Pix2NeRF and HeadNeRF, in Figure~\ref{fig:img2nerf_comparison}. 
Pix2NeRF~\cite{cai2022pix2nerf} proposed an encoder to translate images to the Pi-GAN~\cite{chan2021pigan} latent space and jointly trained the encoder and the Pi-GAN generator. In contrast to Pix2NeRF and our \textit{NeRFInvertor}, which utilize GAN priors for novel view synthesis, HeadNeRF~\cite{hong2022headnerf} proposed a NeRF-based parametric head model to synthesize images with various poses.
The results in Figure~\ref{fig:img2nerf_comparison} show that Pix2NeRF is not a good candidate for real face synthesis, since it does not preserve the identity properly. HeadNeRF also shows a noticeable identity gap between the generated and input images. Furthermore, it also inherits the disadvantage of 3D parametric models that it produces some artificial visuals.
Our approach achieves significantly higher fidelity and better identity preservation compared to these two methods.

\noindent\textbf{Evaluation on multiple NeRF-GANs.}
We validate our pipeline on a variety of NeRF-GANs, including AniFaceGAN~\cite{wu2022anifacegan}, GRAM~\cite{deng2022gram}, and EG3D~\cite{chan2022eg3d}.
In Figure~\ref{fig:novel_views_from_multi_GAN}, we show the reconstruction, novel views, and expressions synthesis given a single input image.
AniFaceGAN uses a deformable NeRF structure and is trained for a dynamic scene. We demonstrate that \textit{NeRFInvertor} can faithfully translate a single image to a deformable NeRF representation, allowing us to generate realistic face editing with controllable poses and expressions (Figure~\ref{fig:showcase} and \ref{fig:novel_views_from_multi_GAN}). 
Given a fixed latent code, GRAM and EG3D generate the NeRF representation for a static scene. We also validate our method on these two methods, demonstrating that we can invert a single image to a traditional NeRF representation for static scenes.

\begin{figure}[t]
  \centering
  \includegraphics[width=\linewidth]{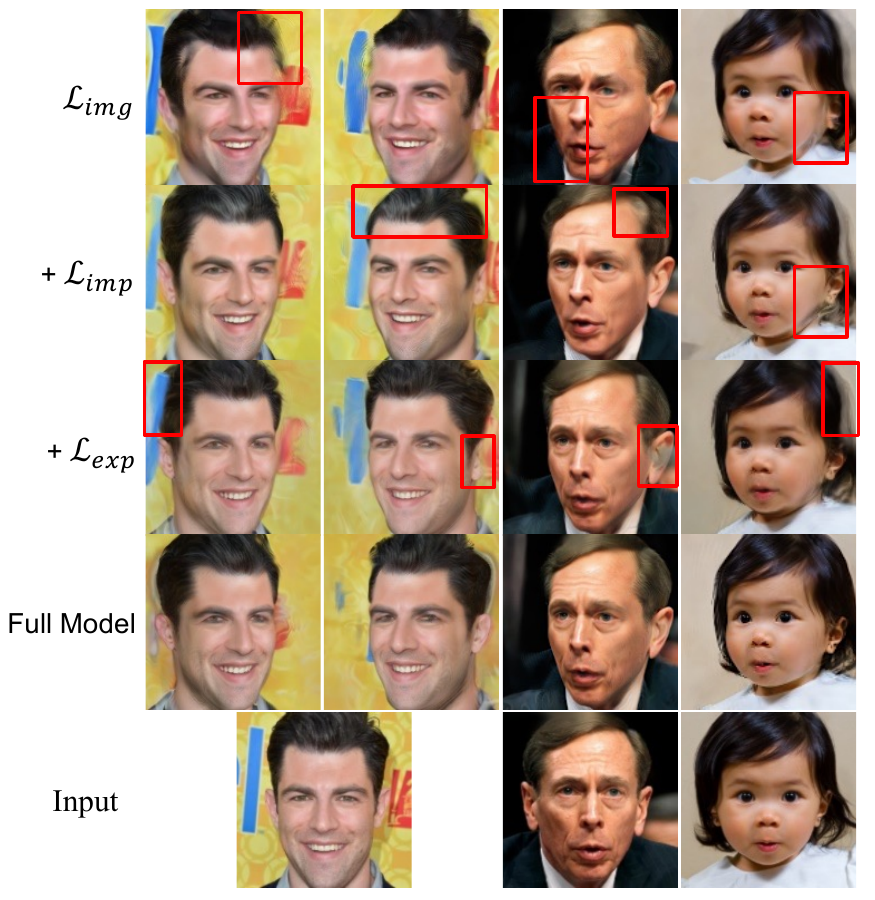}
   \caption{\textbf{Ablation study on different regularizations.} 
   }
   \label{fig:ablation}
\end{figure}

\begin{table}
    \centering
    \caption{\textbf{Quantitative results of different regularizations.} }
    {
    \begin{adjustbox}{max width=\linewidth}
        \begin{tabular}{cccccc}
            \toprule
             & \multicolumn{1}{c}{Rec.} & \multicolumn{2}{c}{Novel View} & \multicolumn{2}{c}{Novel View+Exp.}\\
             & PSNR/SSIM ($\uparrow$) & FID ($\downarrow$) & ID ($\uparrow$) & FID ($\downarrow$) & ID ($\uparrow$)\\\midrule 
            $\mathcal{L}_{img}$ & 22.48 / 0.764 & 38.61 & 0.66 & 38.82 & 0.60\\
            $+\mathcal{L}_{imp}$  & 22.61 / 0.761 & 37.83 & 0.73 & 36.99 & 0.66 \\
            $+\mathcal{L}_{exp}$  & \textbf{25.64} / 0.828 & 34.68 & 0.76 & 34.55 & 0.66 \\
             \rowcolor{lightgray}
            Full Model  & 25.61 / \textbf{0.830} & \textbf{34.07} & \textbf{0.77} & \textbf{33.61} & \textbf{0.67} \\
            \bottomrule
        \end{tabular}
    \label{tab:ablation}
    \end{adjustbox}
    }
\end{table}

\subsection{Quantitative Evaluation}
Quantitative experiments were performed on the first 150 samples from the CelebA-HQ test set and FFHQ dataset. For reconstruction results, we report PSNR in Table~\ref{tab:qualitative_comparison} and additional structural similarity (\ie SSIM~\cite{wang2004image}) and identity similarity in the Supplementary Material. Since we do not have multi-view ground truth images, we report Frechet Inception Distance (\ie FID) and identity similarity for novel view images following prior works~\cite{hong2022headnerf,cai2022pix2nerf}. 
As can be seen, the results align with our qualitative evaluation. I2S ($\mathcal{W}$) fails to preserve the identity of the input image and has poor ID scores. I2S ($\mathcal{W+}$) introduces artifacts in novel views and results in high FID scores. Compared to PTI, our method generate comparable reconstructions, but superior novel view synthesis in terms of visual quality (\ie FID scores) and better identity preservation (\ie ID scores). We achieve the best overall performance across all metrics.
More quantitative results (\eg novel view \& expression evaluations) can be found in the Supplementary Material.

\begin{figure}[t]
  \centering
  \includegraphics[width=0.85\linewidth]{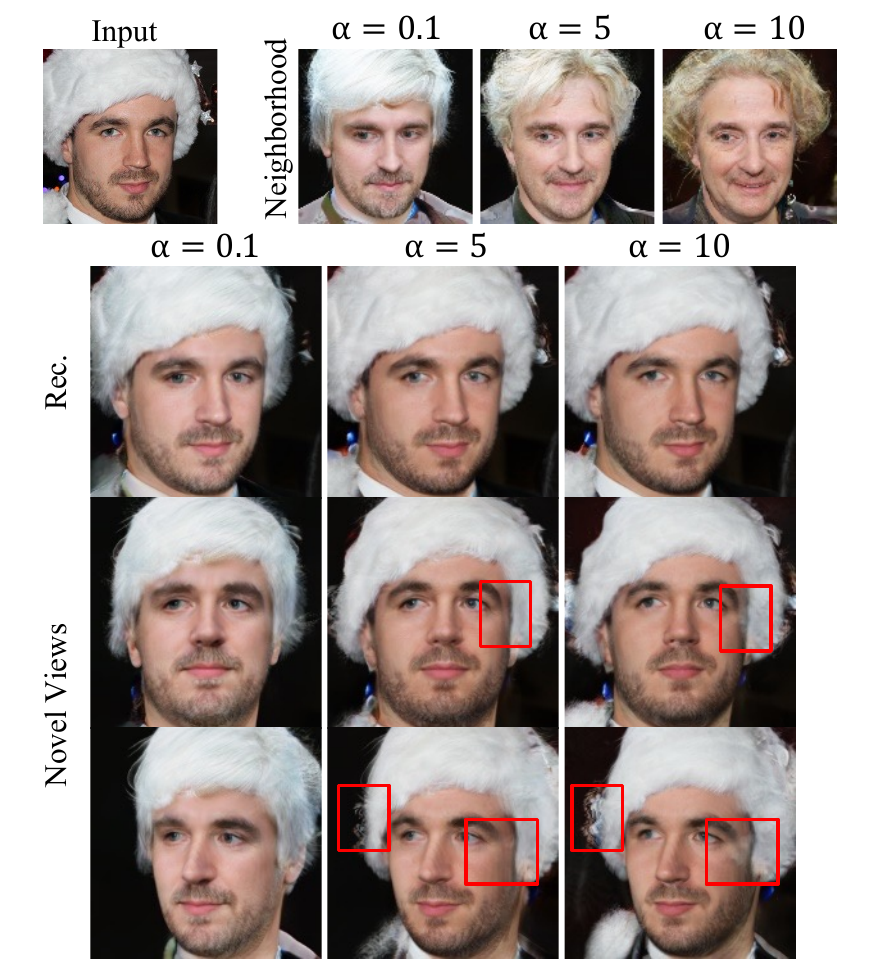}
   \caption{\textbf{Ablation study on distance of neighborhood samples.}}
   \label{fig:ablation_neighbor}
\end{figure}

\subsection{Ablation Study}

\noindent\textbf{Effectiveness of Regularization.}
We conduct an ablation study on the CelebA-HQ test set.
Compared to existing image space losses $\mathcal{L}_{img}$, we show the effects of our proposed implicit geometrical regularization (\ie $\mathcal{L}_{imp}$), explicit geometrical regularization (\ie $\mathcal{L}_{exp}$), and masked regularizations (\ie Full Model) in Figure~\ref{fig:ablation} and Table~\ref{tab:ablation}. The results in Figure~\ref{fig:ablation} indicate that the fine-tuned model with just image space losses $\mathcal{L}_{img}$ is prone to generating artifacts in novel-view images and inaccurate 3D geometry; the implicit implicit geometrical regularization $\mathcal{L}_{imp}$ helps to eliminate artifacts; the explicit geometrical regularization  $\mathcal{L}_{exp}$ improves visual quality and the subject's 3D geometry; the full model with masked regularizations reduces fogging around the hair, ear, or cheek.
The quantitative results in Table~\ref{tab:ablation} are consistent with the qualitative results in Figure~\ref{fig:ablation}.

\noindent\textbf{Neighborhood Selection.}
We empirically find out that the distance between the optimized and neighborhood latent codes affects the ID-preserving ability and geometrical constraints. As shown in Figure~\ref{fig:ablation_neighbor}, if the distance is too small (\ie $\alpha<=1$), the model shows better geometrical constraints but worse identity preservation. And if the distance is too large (i.e., $\alpha>=10$), the model has better ID-preserving ability but undesired 3D geometry with visual artifacts in novel views. 
Setting the distance $\alpha$ within the range $[1,10]$, we are able to well balance the image space supervision and regularization. We therefore simply set the distance to 5 for all the training samples in the experiments.


\section{Conclusion}
   We introduced \textit{NeRFInvertor} as a universal method for a single-shot inversion of real images on both static and dynamic NeRF-GAN models. We employed image space supervision to fine-tune NeRF-GANs generator for reducing identity gap, along with explicit and implicit geometrical constraints for removing artifacts from geometry and rendered images in novel views and expressions. Our experiments validate the importance of each component in our method for 3D consistent, ID-preserving, and high-fidelity animation of real face images.

{\small
\bibliographystyle{ieee_fullname}
\bibliography{egbib}
}

\end{document}